\documentclass{article}
\usepackage{arxiv}

\usepackage[utf8]{inputenc} % allow utf-8 input
\usepackage[T1]{fontenc}    % use 8-bit T1 fonts
\usepackage{hyperref}       % hyperlinks
\usepackage{url}            % simple URL typesetting
\usepackage{booktabs}       % professional-quality tables
\usepackage{amsfonts}       % blackboard math symbols
\usepackage{nicefrac}       % compact symbols for 1/2, etc.
\usepackage{microtype}      % microtypography
\usepackage{graphicx}
\usepackage{color}
\usepackage{xspace}
\usepackage{eurosym}
\usepackage{array,ragged2e}
\usepackage[round]{natbib}
\usepackage{amssymb}
\title{A Multi-Source Entity-Level Sentiment Corpus for the Financial Domain: The FinLin Corpus}

\author{
  Tobias Daudert \\
  Insight Centre for Data Analytics \\
  National University of Ireland, Galway \\
  \texttt{tobias.daudert@insight-centre.org} \\
}

\begin{document}

\maketitle

\begin{abstract}
We introduce FinLin, a novel corpus containing investor reports, company reports, news articles, and microblogs from StockTwits, targeting multiple entities stemming from the automobile industry and covering a 3-month period. 
FinLin was annotated with a sentiment score and a relevance score in the range [-1.0, 1.0] and [0.0, 1.0], respectively. The annotations also include the text spans selected for the sentiment, thus, providing additional insight into the annotators' reasoning.
Overall, FinLin aims to complement the current knowledge by providing a novel and publicly available financial sentiment corpus and to foster research on the topic of financial sentiment analysis and potential applications in behavioural science.
\end{abstract}

% keywords can be removed
%\keywords{Corpus \and Annotation \and Sentiment \and Finance \and Microblogs \and News \and Reports \and Multi-source \and StockTwits \and Automobile}

\section{Introduction}
\label{sec:intro}
Modern societies and their welfare rely on market economies making millions of people around the globe affected by changes in the markets \citep{nassirtoussi2014text}. It is fundamental to understand what influences the markets, particularly the financial markets as a proxy for the market economies, as well as how they react and how strong is this influence.
Within the last 20 years, sentiment analysis aiming at measuring public mood in respect of the financial domain grew to an important field of research. Sentiments are extracted from various data sources, one example is news articles. Within this source, one can find discussions regarding macroeconomic factors, company-specific reports or political information, all of which can be relevant to the markets \citep{sinha2014using}. 
As most of the current research is based on specific news such as financial or corporate, it can be fruitful to combine information from textual sentiment stemming from other publicly available sources such as blogs and company reports. Only a few researchers \citep{gao2009advances,hagenau2013automated,rachlin2007admiral,schumaker2012evaluating,schumaker2009textual} have experimented with such a hybrid approach. However, a dataset combining multiple sources while covering the same period for financial textual sentiment is absent. To complement the current knowledge, we present FinLin, a novel and publicly available financial sentiment corpus consisting of four contemporaneous data types: news, microblogs, company reports, and investor reports. While consisting of multiple data types, it covers the same period and targets the same set of entities across all sources, thus, linking records through a set of shared entities.

%%%%%%%%%%%%%%%%%%%%%%%%%%%%%%%%%%%%%%%%%%%%%%
%%%%%%%%%%%%%%%%% NEW SECTION %%%%%%%%%%%%%%%%
%%%%%%%%%%%%%%%%%%%%%%%%%%%%%%%%%%%%%%%%%%%%%%
\section{Related Work}
\label{sec:background}
Sentiment analysis is currently a trending research topic in natural language processing (NLP). While its technical goal is to analyse peoples' sentiments towards entities and their attributes in written text, its applications are widespread and are already affecting our daily lives; the detection of fake reviews, the study/survey of peoples' political opinion, or automatic stock trading based in Twitter data, are a few examples \citep{liu2012sentiment}. Although the term was only coined in 2003 \citep{nasukawa2003sentiment,liu2012sentiment}, research on sentiment analysis has been very active since 2000 \citep{wiebe2000learning,liu2012sentiment}. Opinion mining is commonly used synonymous to sentiment analysis, however, it is still debatable whether these are distinct terms. Under the umbrella of opinion mining, even more tasks are attributed to this field such as emotion mining, suggestion mining, or subjectivity analysis.
\par
%run-in, tick off, tosser, boob, duff, frog
Initial works on sentiment analysis mainly relied on lexicons \citep{hu2004mining,kim2004determining,nigam2004towards,ding2008holistic}, such as WordNet, while currently, machine learning (ML) and deep learning (DL) techniques are the preferred approach \citep{do2019deep}. While semi-supervised approaches also exist, ML and DL approaches can be broadly classified into supervised, and unsupervised, with the first requiring manually labelled training data which can be labour intensive and costly to obtain. Unsupervised tasks offer an inexpensive alternative, nonetheless, it comes with some caveats, namely the need for large-scale datasets which is difficult to acquire, especially when dealing with multiple data sources. Hence, this leads to the growing need of labelled datasets for sentiment analysis adapted to different domains and environments. It is also important to highlight the necessity of data covering a variety of languages; it has to be accounted for that labels might not be accurate in different settings. For example, \textit{lush} can describe an area with ``a lot of green, healthy plants'' in American English while in British English, it can also mean ``a person who regularly drinks too much alcohol'', although sharing the same language \citep{dictionary_lush}. Semantic change \citep{grzega2007english} plays its role, for example, \textit{awful} originated as a positive term while currently it is used with a negative connotation \citep{wijaya2011understanding}.
\par
Given the many challenges in sentiment analysis, it is natural that a myriad of corpora exist. As such, we highlight a selection of five recent corpora: 
\begin{itemize}
\item \textbf{IMDB Corpus}: Contains 25,000 textual movie reviews from IMDB\footnote{\url{https://www.imdb.com/}} annotated as positive and negative \citep{maas2011learning}.

\vspace{0.3cm}
\item \textbf{Sentiment Stanford Sentiment Treebank}: Comprises reviews from the entertainment review website Rotten Tomatoes\footnote{\url{https://www.rottentomatoes.com/}} annotated using a sentiment scale between 1 and 25 \citep{socher2013recursive}.

\vspace{0.3cm}
\item \textbf{Sentiment140}: Consists of 160,000 tweets annotated in polarities (positive, neutral, negative) after removing emoticons \citep{go2009twitter}.

\vspace{0.3cm}
\item \textbf{SemEval2017 Task 4}: One sub-task deals with polarity detection in tweets in English and Arabic, while another deals with a 5-point scale sentiment classification \citep{rosenthal2019semeval}.

\vspace{0.3cm}
\item\textbf{SemEval2017 Task 5}: Addresses fine-grained sentiment analysis for microblogs (Twitter and StockTwits) and news annotated with a fine-grained sentiment scale between -1 and 1 \citep{cortis-etal-2017-semeval}. 
\end{itemize}
Overall, corpora can consider one type (\emph{e.g.} microblogs) or one source of data (\emph{e.g.} Twitter), as well as being open-domain (\emph{e.g.} Sentiment140) or domain-specific (\emph{e.g.} SemEval2017 Task 5). 
In the case of the financial domain, previous works have focused on stock price prediction and on unsupervised sentiment analysis. \citet{bollen2011twitter} uses Twitter microblogs together with the OpinionFinder sentiment lexicon\footnote{\url{https://mpqa.cs.pitt.edu/opinionfinder/}} to measure correlations between mood in tweets and the stock prices. In a similar fashion, \citet{lee2014importance} aim at the stock price prediction using 8\-k reports, mandatory for U.S. publicly listed companies, together with SentiWordNet\footnote{\url{http://sentiwordnet.isti.cnr.it/}}. \citet{li2014news} use 5 years of FINET\footnote{\url{http://www.finet.hk/mainsite/index.htm}} stock news in the English language and apply the McDonald financial sentiment dictionary \citep{loughran2011liability}, as well as the Harvard IV-4 sentiment dictionary to identify sentiments\footnote{\url{http://www.wjh.harvard.edu/~inquirer/homecat.htm}}.
The SemEval 2017 Task 5 was the first labelled sentiment analysis corpus dealing with fine-grained sentiment, in the financial domain. It consisted of two sub-tasks, the first was focused on microblogs while the second dealt only with news data. However, the microblogs and news do not cover the same period nor entities, thus, its use is limited as it cannot be applied in tasks requiring contemporaneous data. This issue was faced in our recent work \citep{daudert-etal-2018-leveraging,daudert-buitelaar-2018-linking}; to obtain news contemporaneous to the microblog data, we faced several limitations as the news collection was performed almost two years after the publication of the SemEval dataset and resulted in a medium-sized dataset.
\par
With the identified gaps in mind, as well as the requirements needed for further advancements in the field, we develop and release the novel FinLin dataset. This labelled dataset covers multiple data sources and types (\emph{i.e.} microblogs, news, investor, and company reports), the same period, and the same entities.
%%%%%%%%%%%%%%%%%%%%%%%%%%%%%%%%%%%%%%%%%%%%%%
%%%%%%%%%%%%%%%%% NEW SECTION %%%%%%%%%%%%%%%%
%%%%%%%%%%%%%%%%%%%%%%%%%%%%%%%%%%%%%%%%%%%%%%
\section{Corpus Design}
\label{sec:corpusdesign}
The aim behind FinLin is to enable the exploration of sentiments directed at targets across different data sources in a financial setting. Throughout this paper, we utilise the terms \textit{target}, \textit{entity}, and \textit{company} interchangeably to represent the sentiment target \emph{i.e.} the companies defined in section \ref{sec:entities}. 
\par
To select the data sources covered by FinLin, we rely on past research to choose four distinct sources representing three data types: microblogs, news articles, and reports. While \textit{data source} refers to the origin of the data, \textit{data type} classifies it into general categories based on their source and attributes.
Based on the only available labelled sentiment dataset for finance \citep{cortis-etal-2017-semeval}, we include StockTwits as the microblog source. Our previous works \citep{daudert-etal-2018-leveraging,daudert-buitelaar-2018-linking} indicated a significant relation between news and stocktwits, hence, we also consider news articles. Besides, we include analyst reports and company reports to capture a professional view and first-hand information on a company. To ensure date-time compatibility, the data collection occurred simultaneously on all sources during the year 2018. To ensure matching information, we defined a set of targets stemming from the automobile sector, beforehand. The data was then annotated and consolidated by domain experts. The following sections detail the sources and entities considered, the information collected, as well as the annotation process.

\subsection{Entities}
\label{sec:entities}
For this dataset, we consider data regarding the automobile sector as it is currently facing a phase of changes (\emph{e.g.} electric cars, $CO_2$ emissions regulations, self-driving cars) with several stakeholders (\emph{e.g.} Volkswagen, Toyota, Ford), hence, receiving wide coverage. We initially limit the data collection to the worldwide top 20 car manufacturers based on the number of produced vehicles as published by the \textit{Organisation Internationale des Constructeurs d'Automobiles} (OICA) \footnote{\url{http://www.oica.net/wp-content/uploads/World-Ranking-of-Manufacturers-1.pdf}}.
\begin{table}[!ht]
\caption{Automobile companies, their nicknames, cashtags, ticker, as well as subsidiaries and their brands. Subsidiary companies in brackets were not considered due to their high ambiguity.}
\begin{tabular}{>{\RaggedRight\arraybackslash}p{4.5cm}|p{1.4cm}|p{1cm}|p{1.2cm}|>{\RaggedRight\arraybackslash}p{5.5cm}}
\toprule
\multicolumn{1}{c|}{\textbf{Company Name}} & \multicolumn{1}{c|}{\textbf{Nickname}} & \multicolumn{1}{c|}{\textbf{Cashtags}} & \multicolumn{1}{c|}{\textbf{Ticker}} & \multicolumn{1}{c}{\textbf{Other subsidiaries}} \\
\bottomrule
\midrule
Toyota Motor Corp & Toyota & \$TM & TM & Hino, Lexus \\ \midrule
Volkswagen AG & VW & \$VLKAY & VLKAF VLKAY POAHF & Audi, Bentley, Bugatti, Ducati, Lamborghini, MAN, Porsche, Scania, (Seat), Skoda \\ \midrule
Hyundai Motor Co & Hyundai & --- & HYMLF HYMTF KIMTF & Genesis, Kia \\ \midrule
General Motors Company & G.M. & \$GM & GM & Buick, Chevrolet, Holden \\ \midrule
Ford Motor Co & Ford & \$FORD & F & Cadillac, Lincoln \\ \midrule
Nissan Motor Co & Nissan & \$NSANY & NSANY & Datsun, (Infiniti) \\ \midrule
Honda Motor Co & Honda & \$HMC & HMC & Acura\\ \midrule
Fiat Chrysler Automobiles NV & Fiat & \$FCAU \$RACE \$FERI & FCAU & Abarth, Alfa Romeo, Chrysler, Dodge, Lancia, Maserati \\ \midrule
Renault SA & Renault & --- & RNSDF, RNSLY & Dacia, Lada \\ \midrule
Peugeot SA & Peugeot & \$UG & PEUGF & Citro\''{e}n, Opel, Vauxhall\\ \midrule
Suzuki Motor Corp & Suzuki & \$SZKMY & SZKMY & --- \\ \midrule
Navistar & --- & \$NAV & NAV & ---\\ \midrule
Daimler AG & Daimler, Mercedes, Mercedes-Benz & \$DDAIF & DDAIF & Smart \\ \midrule
Bayerische Motoren Werke AG & B.M.W. & \$BMW & BMWYY & Mini Cooper \\ \midrule
\bottomrule
\end{tabular}
\label{table:entities}
\end{table}
Given our focus on the English language, a portion of the mentioned companies have a reduced representation or are absent in the four distinct sources considered. Thus, to avoid data sparseness of certain entities in FinLin we excluded the following companies: SAIC Motor Corporation Limited, Suzuki K.K., Geely Automobile Holdings Ltd, Chongqing Changan Automobile Co Ltd, Mazda Motor Corporation, Dongfeng Motor Group Co Ltd, BAIC Motor Corporation Ltd, Mitsubishi Motors Corporation. To track the remaining, we also shortlisted the respective brand names and subsidiary companies; the final list of selected companies is shown in table \ref{table:entities}. Texts referring to Porsche and Kia were also collected using the tickers POAHF and KIMTF, respectively; to present the results we aggregate these accordingly to the parent company \emph{i.e.} Volkswagen and Hyundai. 

\subsection{Sources}
\label{sec:sources}
\subsubsection{Microblogs}
\label{sec:microblogs}
The rapid expansion of social media has changed how people communicate and express their opinions making it one of the main sources of public sentiment. Founded in 2008, StockTwits\footnote{\url{https://stocktwits.com/}} is a microblog platform tailored to investors, with over 2 million users currently registered\footnote{\url{https://about.stocktwits.com/}}. In a similar fashion to Twitter, users share short messages referred to as stocktwit(s). However, a particular characteristic of a stocktwit is the addition of \textit{cashtags}. A cashtag corresponds to any Stock, Future, or Forex ticker prefixed with the \$ symbol; for example, Toyota Motor Corporation is represented by the \$TM cashtag. This enables the aggregation of all information targeting a determined target.
The collection of stocktwits was performed using the StockTwit Application Programming Interface (API)\footnote{\url{https://api.stocktwits.com/developers/docs}} utilising the selected company names and respective chastags.

\subsubsection{News}
\label{sec:news}
News articles, written work which is published in either a print or electronic medium, 
%have been shown an important proxy for stock price prediction and the financial markets
have been shown important for stock price prediction and the financial markets by proxy. However, as the amount of news articles is continuously increasing, machine-based systems are crucial to filter noise \citep{li2014news}. News articles can also be long dealing with one or many events which are put into context.
News were retrieved from Yahoo! News\footnote{\url{https://news.yahoo.com/}}; as it is aggregating articles from multiple providers, this platform provides a simple and convenient way to cover several newspapers. Yahoo! News was accessed daily to find and store all news articles containing at least one of the terms shown in table \ref{table:entities} in the headline.

\subsubsection{Company Reports}
\label{sec:companyreports}
Company reports reflect companies' financial performance and strategy, and typically include quantitative data (accounting and financial data) and qualitative data (narrative texts). As proposed by \cite{hajek2013evaluating}, these reports also describe the managerial priorities of a company, hence, they tend to differ in terms of the subjects emphasized when the company's performance worsens \citep{kohut1992president}.
For this corpus, we also chose to focus on company reports since they provide first-hand information by company officials and play a crucial role when it comes to assessing a company's performance. As there is no automated solution for the retrieval of reports, we manually gathered the publicly available company reports from the respective company's website which usually come in a PDF format.

\subsubsection{Investor Reports}
\label{sec:investorreports}
Investor reports, also called analyst reports, provide information about an analyst's assessment of a company and its future performance. Different from opinions on StockTwits, this type of reports provides detailed reasoning and is often done by professionals who would closely follow a group of companies for multiple years. The investor reports were collected from Seeking Alpha\footnote{\url{https://seekingalpha.com/}}, a crowd-sourced content service for financial markets with investors and industry experts as contributors.

\subsubsection{Timeframe Selection}
\label{sec:timeperiodselection}
\begin{figure}[ht]
    \centering
  \includegraphics[width=\linewidth]{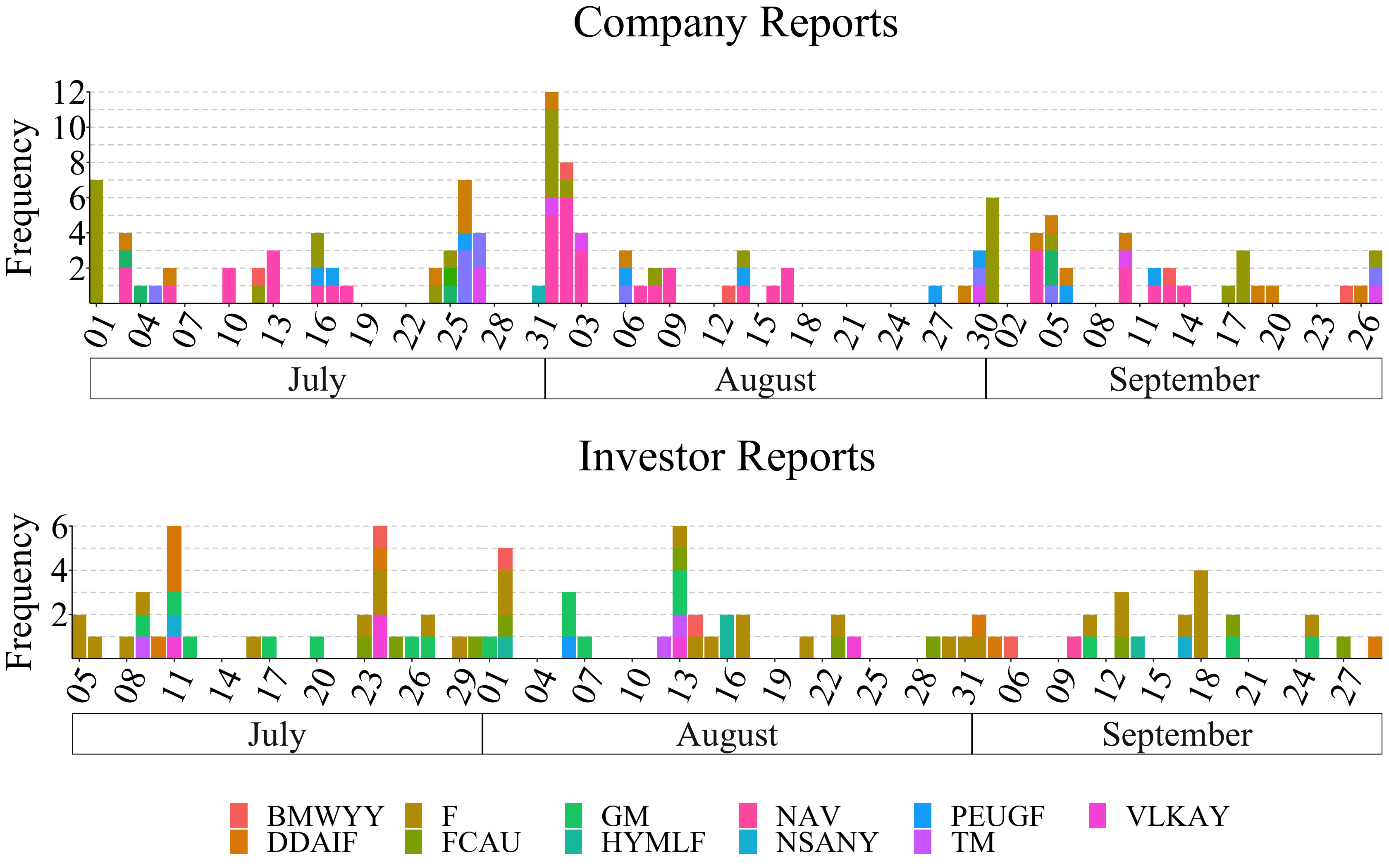}
  \caption{Distribution of all reports per target in FinLin. The reports are aggregated according to the tickers specified in table \ref{table:entities}. The x axis is labelled with a 3-day scale.}
  \label{fig:reports_distribution}
\end{figure}

\begin{figure}[ht]
\centering
  \includegraphics[width=0.94\linewidth]{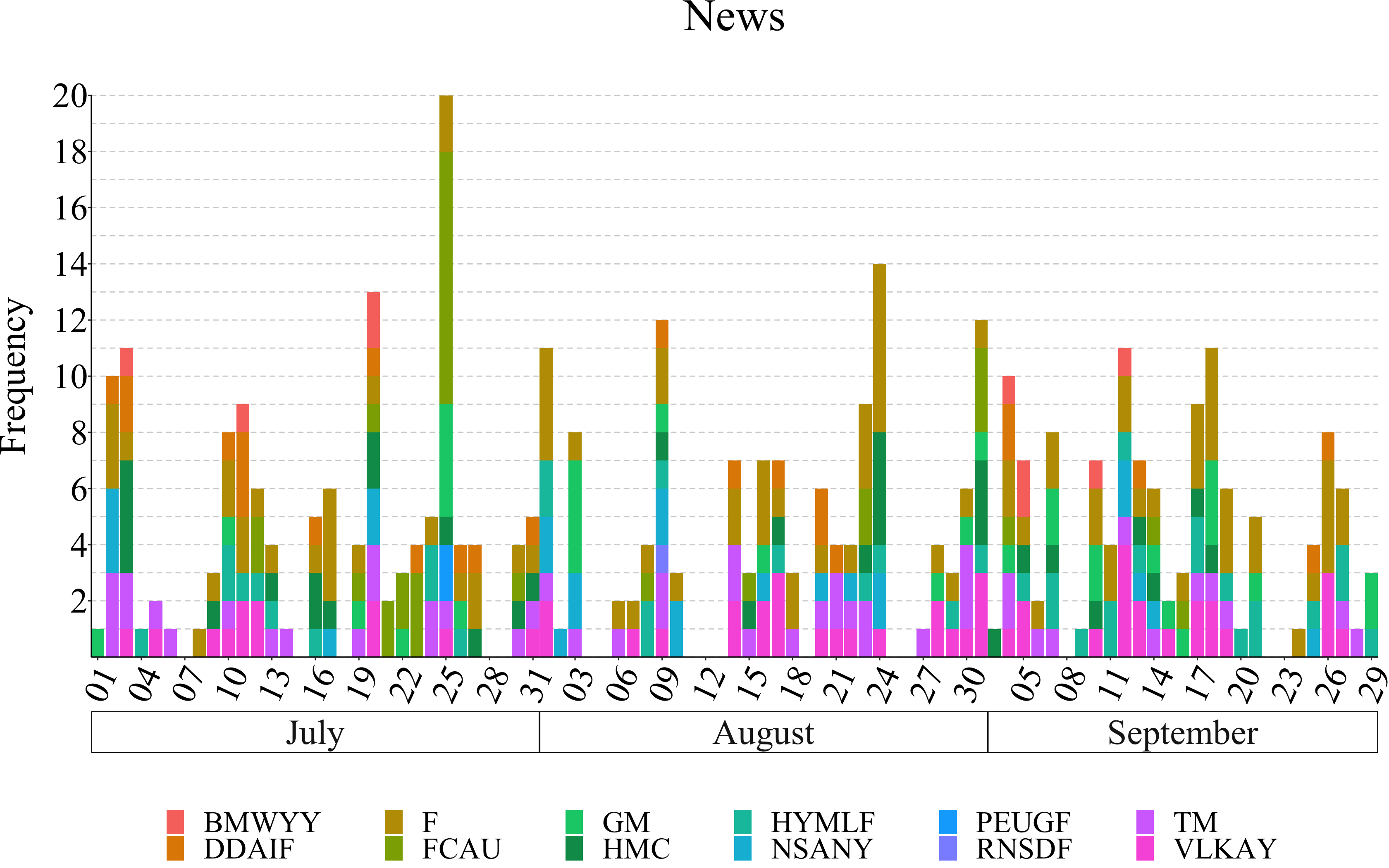}
  \caption{News distribution in FinLin. The news are aggregated based on the tickers present in table \ref{table:entities}. The x axis is labelled with a 3-day scale.}
  \label{fig:news_distribution}
\end{figure}

\begin{figure}[!ht]
\centering
  \includegraphics[width=0.94\linewidth]{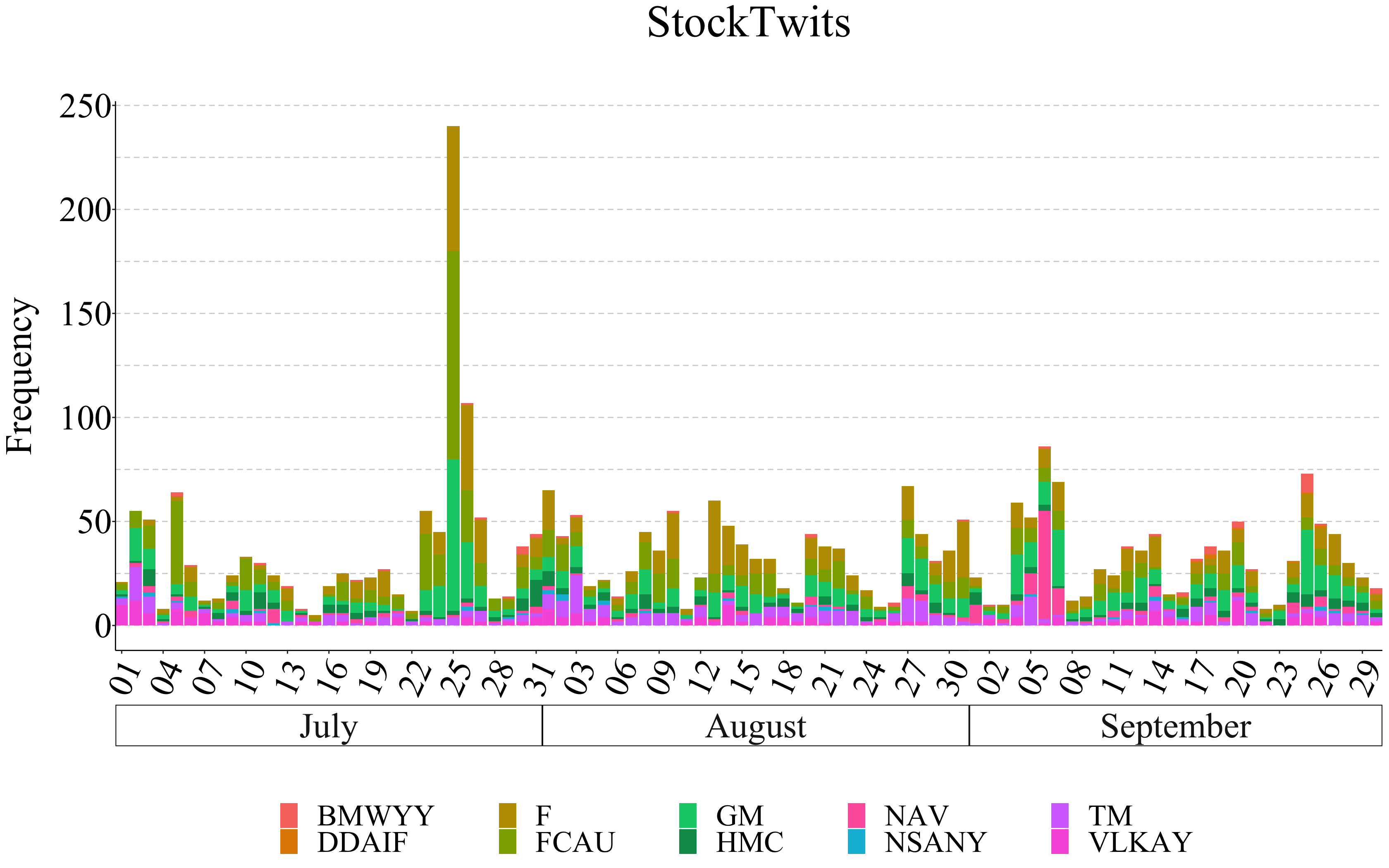}
  \caption{Distribution of the stockwits in FinLin. The stocktwits are aggregated according to the tickers specified in table \ref{table:entities}. The x axis is labelled with a 3-day scale.}
  \label{fig:stocktwits_distribution}
\end{figure}

We initially collected data for the entire year 2018, for all targets and sources. Given the available budget for the data annotation (see section \ref{sec:annotatorselection}), we choose a period of 3 months matching the common standard of quarterly reporting, hence, ensuring the occurrence of company reports within this timeframe. We specifically choose the period of 01/July/2018 to 30/September/2018 based on the data diversity and its volume which is highest in the 3\textsuperscript{rd} quarter of 2018. Figure \ref{fig:reports_distribution} shows the frequency distribution for the investor and company reports; figure \ref{fig:news_distribution} and figure \ref{fig:stocktwits_distribution} present the distribution for the news and stocktwits, respectively. 
%%%%%%%%%%%%%%%%%%%%%%%%%%%%%%%%%%%%%%%%%%%%%%
%%%%%%%%%%%%%%%%% NEW SECTION %%%%%%%%%%%%%%%%
%%%%%%%%%%%%%%%%%%%%%%%%%%%%%%%%%%%%%%%%%%%%%%
\newpage
\section{Corpus Annotation}
\label{sec:corpusannotation}
Given our selected data covering a period of 3 months, the shortlisted entities and their subsidiaries, we conduct the following pre-processing steps in preparation for the annotation task. Specifically, we conduct down-sampling for the stocktwits, sentence extraction for the news articles and investor reports, and a manual text extraction for the company reports.

\subsection{Data Preparation}
\label{sec:datapreparation}
\subsubsection{Stocktwits Down-Sampling}
The initial data collection for the specified period yield 13,243 stocktwits; however, given that 91.7\% of the stocktwits deal with the entities F, GM, and FCAU, we chose to limit the maximum number of stocktwits to 700 per entity to avoid data over-representation and due to monetary constraints. We apply a sampling algorithm, from the Pandas library\footnote{\url{https://pandas.pydata.org/}}, to obtain the final dataset consisting of 3,204 stocktwits. 

\subsubsection{Sentence Extraction for News and Investor Reports}
Given the nature of news articles and investor reports, consisting of longer descriptive and contextualised texts, we decide for the annotation of text portions certainly dealing with the entity of interest.
During initial exploration of the collected news and reports, we noticed a shared behaviour: the articles' core information either appears at the beginning of a section dealing with the tracked entity or in the next few sentences. Some investor reports even deal with multiple entities at the same time, thus, not all of the text is important to our analysis. Hence, we extract the sentence in which the entity occurs first and the following two sentences. The extracted three sentences and the title of the news/report are then subject to the annotation task. Note that there are also cases in which other entities are mentioned in the content but do not constitute the main article topic. Therefore, we focus on annotating entities specifically named in the headline of a news article.

\subsubsection{Text Extraction for Company Reports}
\label{sec:textextraction}
Company reports are usually provided in PDF format, as such, we had to manually extract their content. Although tools for the automatic extraction of PDF content exist, we decided to conduct a manual text extraction based on two reasons: Firstly, as the analysis goal of this dataset is of a qualitative nature, extraction issues could severely affect the results; secondly, as we are dealing with a limited number of reports, small extraction errors could lead to contrasting differences in the final statistics. Furthermore, the dataset's reasonable size allows for undemanding manual extraction.
\begin{figure*}[!ht]
\centering
\includegraphics[scale = 0.09]{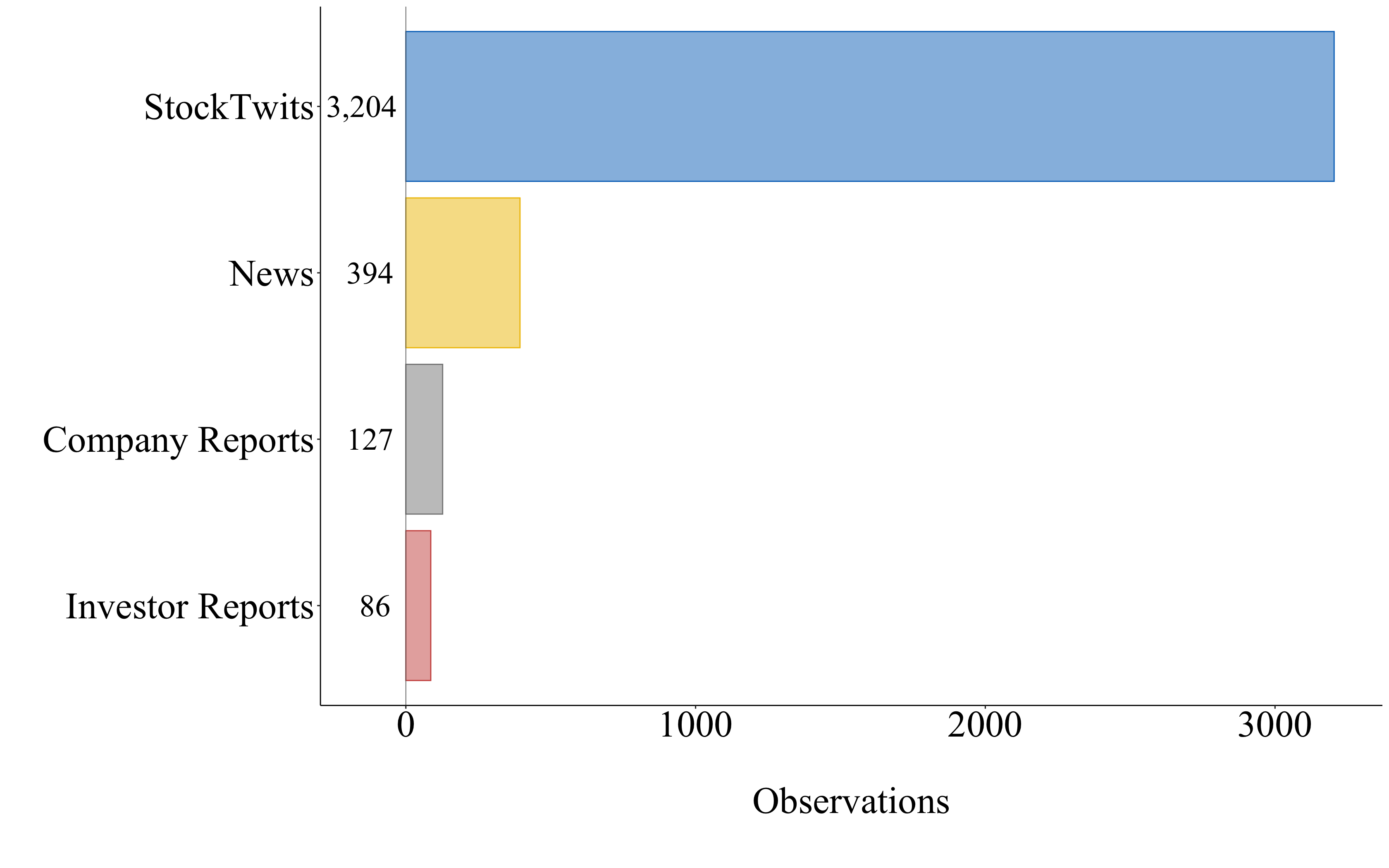}
\caption{Number of observations for each data source.}
\label{fig:observations}
\end{figure*}
\par
\begin{figure*}[!ht]
\centering
\includegraphics[width=0.8\textwidth]{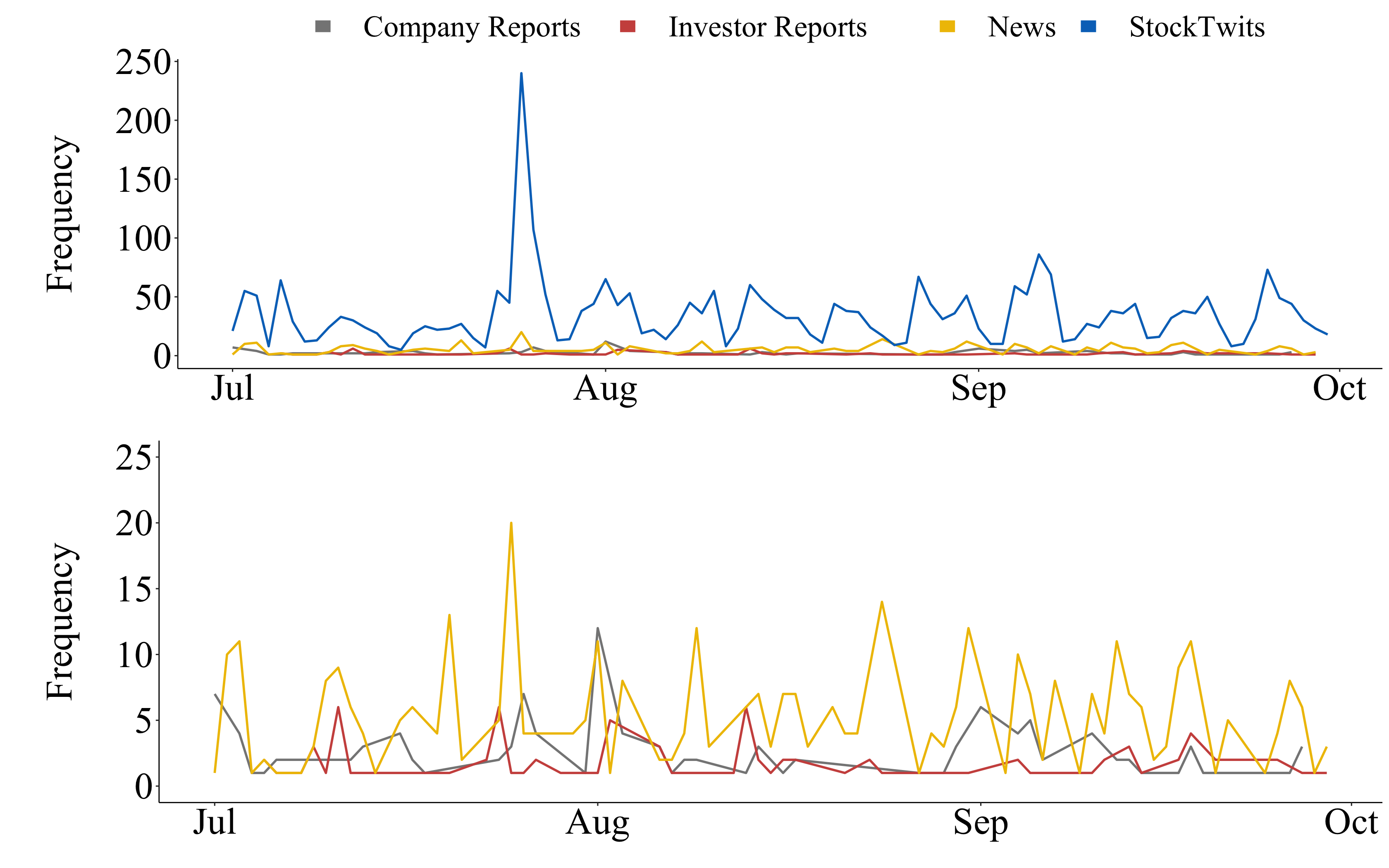}
\caption{Time series distribution for sources in the FinLin corpus. The bottom plot shows the distribution for the Company Reports, Investor Reports, and News, using a smaller y-axis. }
\label{fig:timeseries}
\end{figure*}

The total number of investor reports, company reports, news, and stocktwits are presented in figure \ref{fig:observations} and the respective time-series in figure \ref{fig:timeseries}. Note that, in these figures, we report on the number of entities per news; as previously stated the same news article can be annotated for multiple entities.

\subsection{Point of View and Sentiment Definitions}
One of the main challenges when understanding, interpreting, and annotating texts for sentiments corresponds to the correct \textit{point of view}. This refers to the position an annotator places themselves, in an imaginary context, reflecting their interests. Taking figure \ref{fig:consolidationmode} as an example, the annotator can interpret the text from the point of view of the author ``glad I sold my position [...]'', or from the point of view of Ford (F) (\emph{i.e.} what does the author feel or think about Ford or what does the selling of the author's position implies for Ford). As one of the aims of this dataset is to support analysis of the selected automobile companies, the point of view refers to the presented target. In the previous example, this corresponds to Ford.
\par
Some of the data types included in FinLin, namely the news and the reports, naturally do not contain opinionated information. In this context, sentiment rather refers to an assessment of the presented information given the desired point of view. Therefore, FinLin incorporates \textit{factual} as well as \textit{subjective} sentiments, but always from the point of view of the specified company. A \textit{factual sentiment} is a sentiment based on underlying facts which can occur, for example, in a news article \emph{e.g.} ``the company is expected to increase its sales'' which itself does not reveal any connotation about the author's feelings regarding this information. A \textit{subjective sentiment} is a sentiment which comes with a subjective evaluation of a piece of information, or reveals what an author thinks or feels \emph{e.g.} ``glad I sold my position'', ``I like that the company is expected to increase its sales''. This is a strong difference to the existing sentiment datasets which only annotate the author's (subjective) sentiment.
Applying \cite{liu2012sentiment} definition of sentiment, we formalise a \textit{subjective sentiment} as a quadruple,
\begin{equation}
    (g, s, h, t)
\end{equation}
where $g$ corresponds to the sentiment target, $s$ the sentiment about the target, $h$ the sentiment holder, and $t$ the time of expression. The formal representation of a \textit{factual sentiment} is defined as a quintuple,
\begin{equation}
    (g, f, s, h, t)
\end{equation}
where $f$ corresponds to the sentiment forming fact and $s$ now reflects the sentiment about the target, based on the fact.
\par
Looking at a trading context, positive sentiment is often seen equal to \textit{bullish} while negative sentiment is seen as \textit{bearish} (\emph{i.e.} buy/long or sell/short)\footnote{\url{https://www.investopedia.com/terms/m/marketsentiment.asp}}. Given the point of view of the target entities, this definition is in line with our definition of factual sentiment, as well as subjective sentiment. Figure \ref{fig:consolidationmode} again shows an intricate example: ``glad I sold my position. Bought a few shares at 9.74 for a start. Gonna buy at the bottom this time''. On one hand, the author expresses his subjective sentiment ``glad I [...]'' and on the other hand, they reveal a bearish sentiment ``Gonna buy at the bottom this time'' which implies his future expectation of Ford (F) reaching the bottom (\emph{i.e.} decreasing in value). Based on the point of view (Ford), the sentiment is expressed in the latter extract \emph{i.e.} bearish sentiment.
An investor can have a positive (subjective) sentiment regarding a company if they are invested short meaning they expect a decreasing company performance or stock price while this is not good for the company itself and implies a bearish (trading) sentiment.
Contrary, they can have a negative (subjective) sentiment if invested short on a value-increasing (bullish) stock. Therefore, positiveness and negativeness always depend on the point of view.

\subsection{Annotator Selection}
\label{sec:annotatorselection}
As FinLin refers to a specific domain, we employed a process to find and select annotators with knowledge in finance and stock trading. The process began with the task advertisement in social media, university channels as well as the local School of Business and Economics. In total, 26 people responded to the advertisement; upon surveying the respective \textit{Curriculum Vitae} we selected 12 people given the criteria:
\begin{enumerate}
    \item Higher education in Finance, Business, or Economics;
    \item Professional experience related to Finance, Business, or Economics;
    \item Stock trading experience;
    \item English native.
\end{enumerate}
These potential annotators were then invited to a selection test in which they were tasked to annotate 100 texts. Given the quality of the annotations (reviewed by the author), the time needed for each annotation, and the ability to work with the annotation interface, the final selection corresponded to three annotators and an additional back-up annotator. During the annotation task, each of the three annotators received a University stipulated hourly compensation of 25.09\euro\xspace(23.24\euro\xspace + 8\% holiday pay entitlement), which is above the minimum hourly wage of 9.80\euro\xspace in the Republic of Ireland, in 2019.
\subsection{Annotation Tool}
\label{sec:annotationtoolandprocess}
\begin{figure*}[!ht]
\begin{center}
\fbox{\includegraphics[scale=0.44]{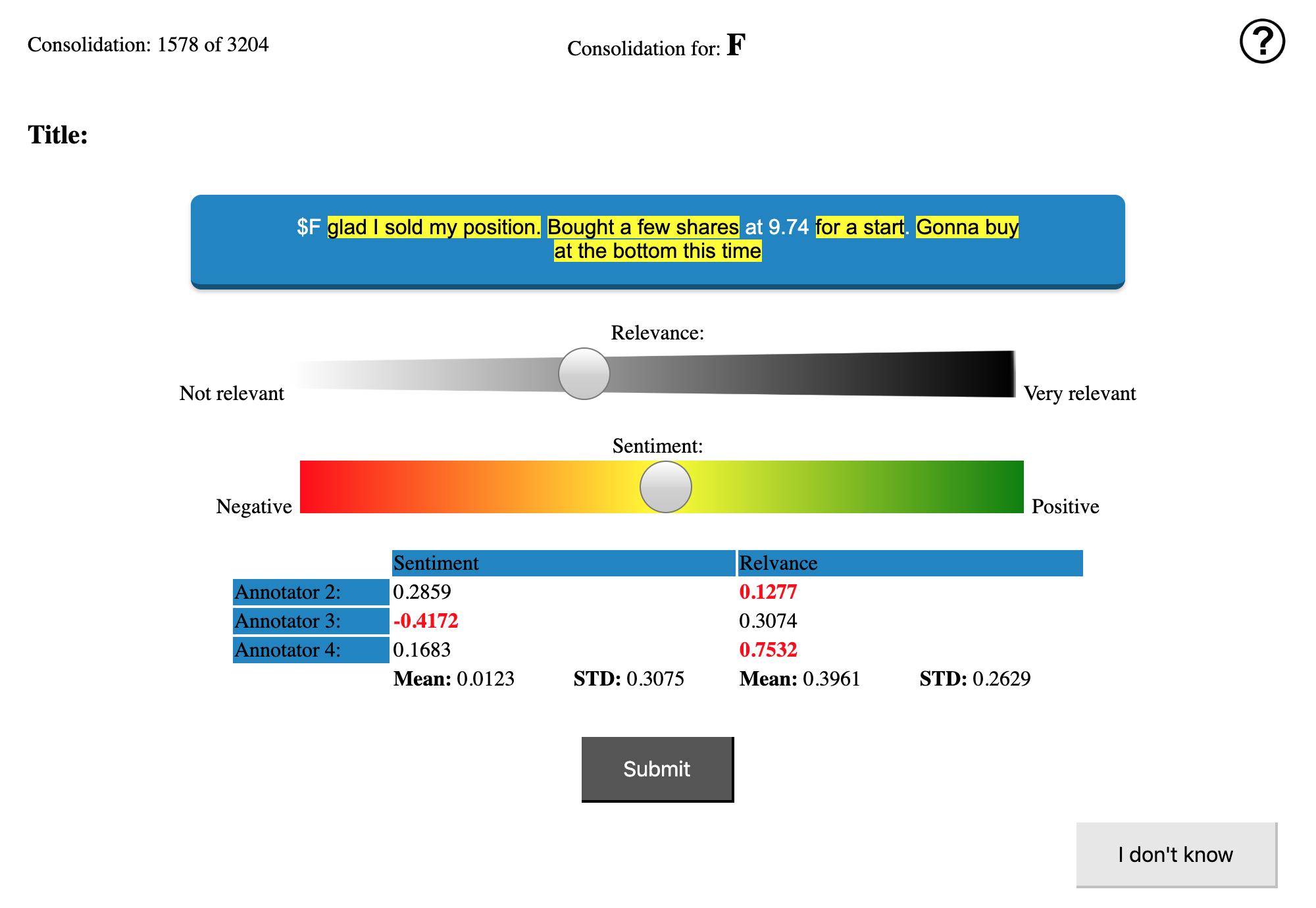}}
\caption{Close-up of the consolidation mode in the applied tool.}
\label{fig:consolidationmode}
\end{center}
\end{figure*}

To aid the annotation process, we utilise AWOCATo \citep{daudert2020a}, a custom-built annotation tool based on CoSACT \citep{daudert-etal-2019-cosact}, the tool utilised to annotate the SemEval Task 5 Financial dataset \citep{cortis-etal-2017-semeval}. It employs a continuous sentiment scale ranging from -1 to 1, with 0 as neutral, as well as a highlight option to select the text portions expressing the respective sentiment. AWOCATo includes a relevancy scale to determine if a given text (\emph{i.e.} microblog, news, or report) is relevant for a determined target. The annotation for the relevance follows the same scheme as for the sentiment, utilising a scale (ranging from 0 to 1). If the annotator is not able to provide an annotation, for example, due to insufficient knowledge, they can select the ``I don't know'' option. In addition, the annotation tool includes a consolidation mode to enable additional input from a consolidator when the discrepancy among annotators for an annotation exceeds a predefined threshold. The interface of this tool in the consolidation mode is shown in figure \ref{fig:consolidationmode}; during the annotation phase, the table above ``Submit'' is not shown, the text is not pre-highlighted, and the scales default is set to 0.
\par
During the annotation period, the selected annotators were invited to attend 10 sessions with a duration of three hours to avoid exhaustion and jeopardising the annotations. To ensure constant concentration levels, the annotators could take breaks or finish the annotation session earlier. All these sessions had the author present to elucidate on any given question. Another 5 sessions occurred at home, without the author's presence, as requested by the annotators and given the flexibility of the online-based annotation tool as accessible from any computer.

\subsection{Annotation Guidelines}
\label{sec:guidelines}
To ensure the annotators correctly understand the annotation task and to guide them throughout the annotation, we created a set of guidelines. These were introduced and recurrently available to the annotators; additional graphical guidance was also provided. Furthermore, these were updated to include the annotators' suggestions. Below we detail, \textit{verbatim}, each segment present in the annotation guideline.
\newpage
\subsubsection{Goals}
\begin{enumerate}
    \item Determine how relevant the text displayed is to a presented company name.
    \item In case it is relevant, determine what is the sentiment of the message.
    \item Mark parts of the text which clearly reflect the sentiment.
\end{enumerate}

\subsubsection{Definitions}
\textit{Relevant} in this context means a text which contains information about or related to the company name given (\emph{e.g.} Ford). Depending on the importance of the mentioned event on the company's business, the information stated in the text can be more or less relevant to the company. In the case of completely non-relevant text, you are asked to just click ``Submit'' without moving the relevance slider. In all other cases, please adjust the slider the more to the right the more relevant the given information is for the presented company.
Examples for the company Ford are:
\begin{enumerate}
    \item ``Ford's profit increase by 20\% in 2017'' $\rightarrow$ relevant
    \item ``Ford is teaming up with Tesla for a new collaboration'' $\rightarrow$ relevant
    \item ``Ford's CEO got injured in a car crash'' $\rightarrow$ relevant
    \item ``Harrison Ford is the best actor of the year'' $\rightarrow$ not relevant
\end{enumerate}
As seen, all relevant examples refer to Ford as a company. However, they are not all referring to the same aspect of its business. Your task will be to annotate with higher relevance, the text which, in your opinion, is more relevant for a company's business.
\par
In the economic context, the \textit{sentiment} of a news text reflects on the company's situation or outlook in the future. Since some authors aim at providing an objective piece of text, you might not be able to figure out whether the author thinks the event/fact they are reporting on is good or bad for a company. Therefore, your task is to judge the presented information. A text's sentiment towards a company, an associated entity, or an event can range from positive (bullish) to negative (bearish) with neutral sentiment in between. This is represented by a slider which is neutral on default. Only move the slider in cases the sentiment is not neutral. Move it the more to the edges of the scale the more a text is positive or negative. A text is more positive (or negative), the more impactful a presented event/fact is for a company, or the stronger the language used by the author. This will fully depend on your interpretation of the given information. You are also asked to relatively judge the sentiment based on information you have already annotated. 
Examples for the company Ford are:
\begin{enumerate}
    \item ``Ford is teaming up with Tesla for a new collaboration''$\rightarrow$ positive
    \item ``Ford's CEO got injured in a car crash'' $\rightarrow$ negative
    \item ``Ford should buy Tesla'' $\rightarrow$ neutral
    \item ``Ford is releasing its new business plan'' $\rightarrow$ neutral
\end{enumerate}
Sometimes, specific parts (\emph{i.e.} span) of the text clearly reflect the message's sentiment; while other parts of the text are less informative. If this is the case in the text displayed, please highlight the identified span with your mouse cursor. In case you highlighted too much text, too little text, or you simply changed your mind, please click on ``Remove spans'' in the top-right corner.
Examples for the company Ford are: ``Ford's \textbf{profit increase by 20\% in 2017}'', ``Ford is \textbf{teaming up with Tesla for a new collaboration}'', ``Ford \textbf{will release a new model} at the end of this year'', ``Ford's \textbf{CEO got injured} in a car crash'', ``Ford's \textbf{business seems to be shaky}'', ``Ford \textbf{should buy Tesla}'', ``Ford \textbf{is releasing its new business plan}''.

\section{Corpus Statistics}
\label{corpusstats}
The FinLin corpus contains a total of 3,811 texts: 3,204 stocktwits, 394 news articles, 127 company reports, and 86 investor reports. Table \ref{table:corpusstats} shows the corpus statistics regarding the characters, words, and the annotated sentiment spans. As shown in figure \ref{fig:observations} and \ref{fig:timeseries}, stocktwits provide the majority of the data followed by news articles, company reports and investor reports. StockTwits emphasises information sharing among its users, which occurs daily on multiple occasions; news also occur daily, thus, these are the major contributors for the FinLin corpus. In contrast, company reports occur quarterly. Investor reports can be published roughly twice-yearly per investor or randomly, focusing on event-based reporting \emph{e.g.} when the company reports a profit warning. Thus, the data distribution corresponded to our expectations.
Besides the frequency and number of observations, figure \ref{fig:timeseries} also shows an irregular cyclic pattern for the stocktwits with a clear peak on July 25. As visible in figure \ref{fig:news_distribution} and \ref{fig:stocktwits_distribution}, the most discussed companies where General Motors (GM) and Fiat-Chrysler (FCAU), thus, the peak in stocktwits can be explained by the United States imposed tariffs on steel which affected GM and the departure of Fiat-Chrysler's CEO.
\par
Overall, Ford, GM, and Fiat-Chrysler are the most covered entities, followed by Toyota Motor (TM) and Volkswagen (VLKAY), as present in table \ref{table:totalobs}. StockTwits is based in the United States of America (USA) where it is a known trading and information-sharing platform. Similarly, Ford, GM, and Fiat-Chrysler are also based in the USA which can indicate why these companies receive more attention than the remaining. The least covered entities are the French companies, Peugeot and Renault. As we collected reports, news, and stocktwits only in English, we believe that the little reporting on these targets, as well as low interest from USA users are the reason for the low coverage.
\begin{table}[!ht]
\centering
\caption{Number of words and characters per text per data source, as well as the word count for the positive, neutral, and negative spans. All columns display average values. The symbol \# represents ``number of''.}
\begin{tabular}{p{3cm}|p{1cm}|p{1cm}|p{2cm}|p{2.3cm}|p{2cm}}
\toprule
\multicolumn{1}{c|}{\textbf{Source}} & \multicolumn{1}{c|}{\textbf{Words}} & \multicolumn{1}{c|}{\textbf{Characters}} & \multicolumn{1}{>{\centering\arraybackslash}m{2cm}|}{\textbf{Positive Span \# Words}}& \multicolumn{1}{>{\centering\arraybackslash}m{2.3cm}|}{\textbf{Negative Span \# Words}}& \multicolumn{1}{>{\centering\arraybackslash}m{2cm}}{\textbf{Neutral Span \# Words}}\\
\bottomrule
\midrule
Company Reports & 401 & 2,016 & 196.77 & 24.84 & 2.24 \\ \midrule
Investor Reports & 2,300 & 12,029 & 22.67 & 20.95 & 2.08 \\ \midrule
News Articles & 711 & 4,037 & 18.34 & 7.17 & 7.75\\ \midrule
Stocktwits & 24 & 109 & 4.39 & 4.01 & 2.48 \\ \midrule
\bottomrule
\end{tabular}
\label{table:corpusstats}
\end{table}
\begin{table}[!ht]
\centering
\caption{Total number of occurrences per entity and source.}
\begin{tabular}{m{1.5cm}|m{1.7cm}|m{1.7cm}|m{1.5cm}|m{1.5cm}|m{1.5cm}}
\toprule
& \multicolumn{1}{>{\centering\arraybackslash}m{1.5cm}|}{\textbf{Company Reports}} & \multicolumn{1}{>{\centering\arraybackslash}m{1.7cm}|}{\textbf{Investor Reports}} & \multicolumn{1}{>{\centering\arraybackslash}m{1.5cm}|}{\textbf{News}} & \multicolumn{1}{>{\centering\arraybackslash}m{1.6cm}|}{\textbf{Stocktwits}} & \multicolumn{1}{>{\centering\arraybackslash}m{1.6cm}}{\textbf{Total}} \\
\bottomrule
\midrule
BMWYY & 5 & 4 & 9 & 55 & 73 \\ \midrule
DDAIF & 16 & 8 & 24 & 7 & 55 \\ \midrule
F & 32 & 32 & 96 & 700 & 860 \\ \midrule
FCAU & 1 & 10 & 31 & 700 & 742 \\ \midrule
GM & 5 & 16 & 32 & 700 & 753 \\ \midrule
HMC & 1 & 0 & 34 & 234 & 269 \\ \midrule
HYMLF & 0 & 4 & 38 & 0 & 42 \\ \midrule
NAV & 9 & 1 & 0 & 200 & 210 \\ \midrule
NSANY & 10 & 2 & 25 & 36 & 73 \\ \midrule
PEUGF & 0 & 1 & 2 & 0 & 3 \\ \midrule
RNSDF & 0 & 0 & 1 & 0 & 1 \\ \midrule
TM & 7 & 3 & 48 & 345 & 403 \\ \midrule
VLKAY & 41 & 5 & 54 & 227 & 327 \\ \midrule \midrule
Total & 127 & 86 & 394 & 3204 & 3811 \\
\midrule
\bottomrule
\end{tabular}
\label{table:totalobs}
\end{table}

\subsection{Annotator Performance}
\begin{figure*}[!ht]
\centering
\includegraphics[scale = 0.1]{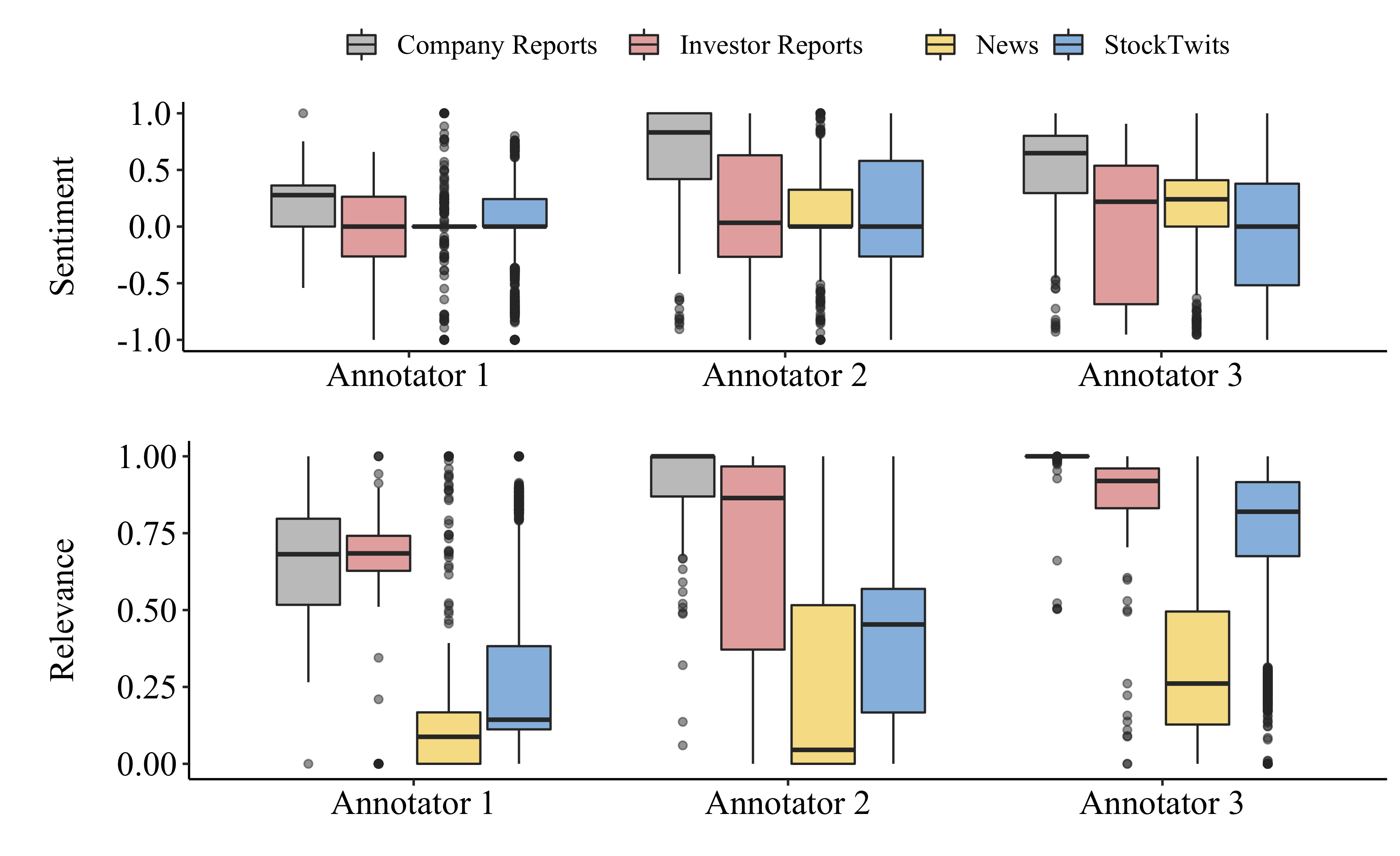}
\caption{Box Plot for the sentiment and relevance annotation per annotator.}
\label{fig:sentiment_relevance}
\end{figure*}

The annotators' performance is reported in figure \ref{fig:sentiment_relevance}. Overall, the sentiment annotation distribution was comparable among the annotators, with the exception of Annotator 1 in the company reports. In contrast, relevance annotation was far more heterogeneous which we attribute to the task difficulty, mainly due to the \textit{point of view} interpretation.
For example, the text ``Mercedes-Benz USA Announces Senior Management Appointments'' had one of the highest discrepancies; annotators 1 and 3 scored the relevance close to zero, and annotator 2 as one. Given that the annotation is performed on the entity Daimler, annotators 1 and 3 might have seen the new appointments, in the USA branch, of less importance to Daimler while annotator 2 potentially interpreted that senior management changes are relevant to the company.
%\subsubsection{Inter-annotator agreement (IAA)}
The inter-annotator agreement (IAA) was calculated utilising Fleiss' Kappa \citep{fleiss1971measuring}. We report on these results for the relevance and sentiment on table \ref{table:IAA}. As the annotation process was performed on a continuous scale, we transformed the sentiment scores into the polarity values (positive $>$ 0, negative $<$ 0, neutral $=$ 0) and the relevance into three categories: low-relevance [0.0,0.25], medium-relevance [0.25,0.75], and high-relevance [0.75,1.0]. 
\par
The IAA for the relevance annotation varies between \textit{slight} for stocktwits and \textit{substantial} for the investor reports according to \cite{landis1977measurement} strength of agreement. 
Comparing the IAA with figure \ref{fig:sentiment_relevance}, annotation trends for each of the three annotators are visible; overall, the distribution (\emph{i.e.} position of the box) of relevance is higher for the investor reports and lower across all annotators for the news, which reach the second-highest IAA. While the IAA for the news is \textit{moderate}, it is \textit{fair} for the company reports.
\begin{table}[ht]
\centering
\caption{Inter-annotator agreement (IAA) for each data source, reported in Fleiss' Kappa.}
\begin{tabular}{m{2cm}|p{1.6cm}|p{1.6cm}|p{1.3cm}|p{1.9cm}|p{1.2cm}}
\toprule
&\multicolumn{1}{>{\centering\arraybackslash}m{3cm}|}{\textbf{Company Reports}} & \multicolumn{1}{>{\centering\arraybackslash}m{3cm}|}{\textbf{Investor Reports}} & \multicolumn{1}{>{\centering\arraybackslash}m{2cm}|}{\textbf{News}} &
\multicolumn{1}{>{\centering\arraybackslash}m{1.9cm}|}{\textbf{Stocktwits}} &\multicolumn{1}{>{\centering\arraybackslash}m{1.2cm}}{\textbf{Overall}} \\
\bottomrule
\midrule
\multicolumn{1}{>{\centering\arraybackslash}m{2.5cm}|}{Relevance IAA} & 0.3298 & 0.6477 & 0.4288 & 0.1219 & 0.1807 \\
\midrule
\multicolumn{1}{>{\centering\arraybackslash}m{2.5cm}|}{Sentiment IAA} & 0.4202 & 0.4663 & 0.3719 & 0.5953 & 0.5610 \\
\midrule
\bottomrule
\end{tabular}
\label{table:IAA}
\end{table}
\par
Comparing the IAA score for the relevance and sentiment, it is clear that the first task poses a higher difficulty than the latter. The IAA is \textit{moderate} for the company reports, and investor reports, \textit{fair-moderate} for the news, and \textit{moderate-substantial} for the stocktwits. We believe this occurs due to the difficulty in defining and applying the point of view. As an example, consider the following text: ``Glad \$F is such a mess! Just keep screwing up new models.'' It is clear what sentiment is portrayed (subjective: positive, factual: negative), however, the relevance relates to the point of view. If the point of view is targeting the author then the text is not relevant (\emph{i.e.} they are a regular person) but if the point of view is targeting the company the text becomes highly relevant (\emph{i.e.} the new car model seems to have issues which affects the future performance of the company). 
In addition, figure \ref{fig:sentiment_relevance} shows that annotator 1 refrains from extreme ratings, contrary to annotator 2 and 3, especially regarding the company reports. Given that we classify the relevance annotations into low, medium, and high, annotator 1 leads to disagreement as there is a tendency towards medium-relevance whereas the others tend to high-relevance. The same is behaviour occurs for the sentiment annotations for which annotator 1 noticeably achieves the smallest distributions.

\subsection{Consolidation}
Having three annotations per text in place, the FinLin dataset was consolidated either automatically or manually by an additional annotator (\emph{i.e.} consolidator). The consolidation was conducted with AWOCATo, the same tool used for annotation. Manual consolidation occurred when (1) at least one of the annotations had a difference higher than the 3\textsuperscript{rd} quantil to the mean of all annotations, or (2) when at least one of the annotators was not able to annotate a given text, or (3) when at least one of the annotations has a different polarity than the remaining. These thresholds are defined in table \ref{table:quantils} as well as the number of automatically and manually consolidated items.
\begin{table}[ht]
\centering
\caption{Thresholds implemented for consolidation, when the value to the mean exceeds the stipulated threshold manual consolidation is used, otherwise the annotations are automatically consolidated. The thresholds were chosen based on the 3\textsuperscript{rd} quantil distribution for the relevance and sentiment annotations.}
\begin{tabular}{m{4cm}|m{3cm}|m{3cm}|m{1.3cm}|m{1.6cm}}
\toprule
&\multicolumn{1}{>{\centering\arraybackslash}m{3cm}|}{\textbf{Company Reports}} & \multicolumn{1}{>{\centering\arraybackslash}m{3cm}|}{\textbf{Investor Reports}} & \multicolumn{1}{>{\centering\arraybackslash}m{1.3cm}|}{\textbf{News}} & \multicolumn{1}{>{\centering\arraybackslash}m{1.6cm}}{\textbf{Stocktwits}} \\
\bottomrule
\midrule
Relevance Threshold & 0.2 & 0.18 & 0.13 & 0.31 \\
\midrule
Sentiment Threshold & 0.31 & 0.3 & 0.21 & 0.26 \\
\midrule
Automatically Consolidated & 51 & 37 & 111 & 1,458 \\
\midrule
Manually Consolidated & 76 \newline (59.84\%)& 49 \newline (56.98\%)& 286 \newline (72.59\%) & 1,746 (54.49\%) \\
%Manually Consolidated & 79 of 131 & 49 of 86 & 267 of 427 & 1746 of 3204 \\
\midrule
\bottomrule
\end{tabular}
\label{table:quantils}
\end{table}
The consolidator's aim was to select the final relevance and sentiment annotation while adhering to the mentioned annotation guidelines. The consolidator did not create an independent label; for a given annotation he utilised the lowest score as lower boundary and the highest score as the upper boundary, thus, serving as a mediator between the initial annotations. However, he also acted as a proofreader for cases in which one or two annotations were misplaced, for example when the text was misinterpreted. As shown in figure \ref{fig:consolidationmode}, the annotator was presented with the initial annotations, the text spans marked according to the annotators' selections, and the scales default set to the mean relevance and mean sentiment scores of the initial annotations.
\begin{figure}[!ht]
\centering
  \includegraphics[width=0.8\linewidth]{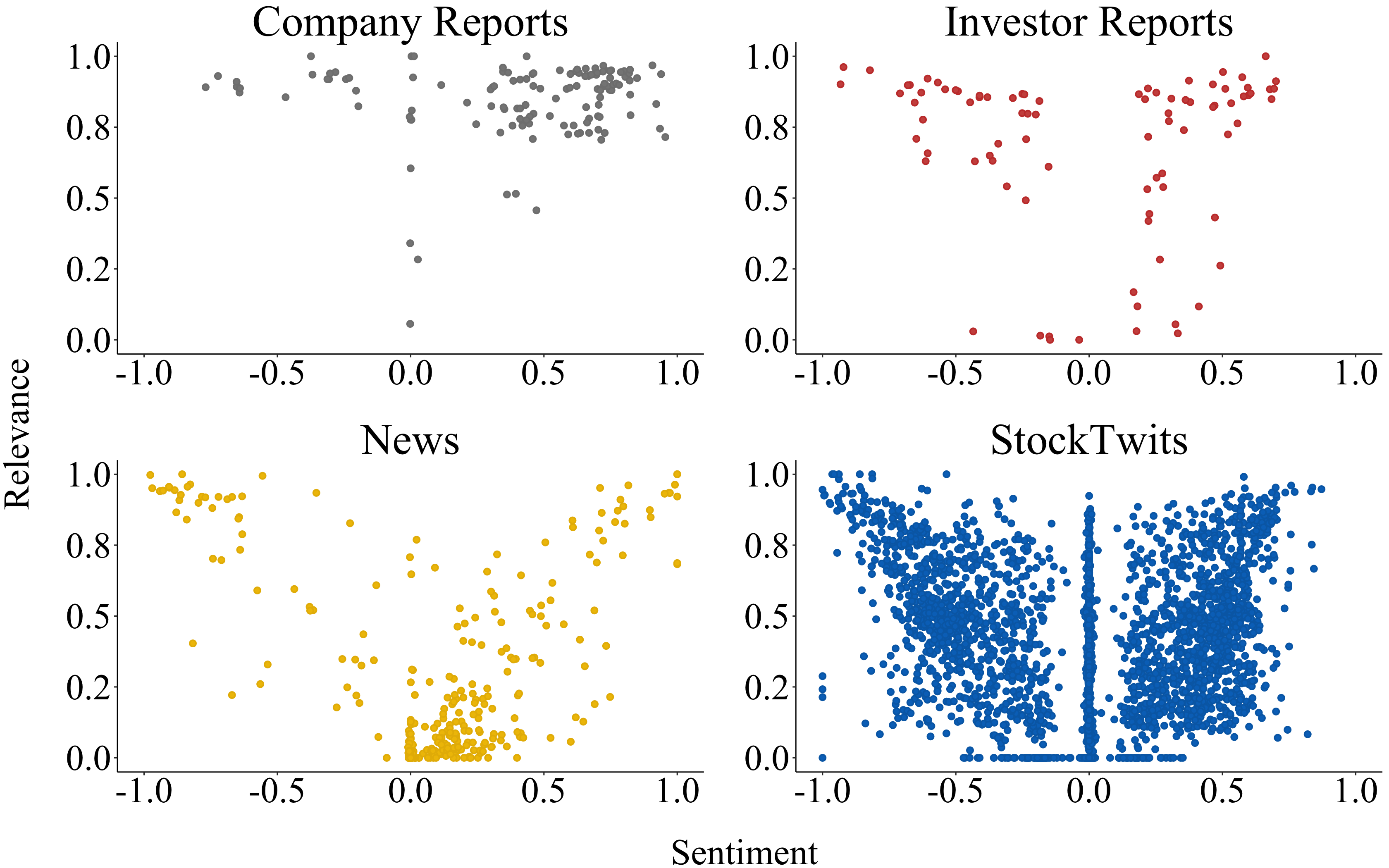}
  \caption{Distribution of the consolidated sentiment and relevance per data source.}
  \label{fig:quad_distribution}
\end{figure}
\par
The final distribution of relevance and sentiment consolidations is shown in figure \ref{fig:quad_distribution}. The majority of the investor reports and company reports tend towards high relevance while news tend to have a lower relevance. In contrast, the stocktwits distribution is sparse, no clear tendency is observed. On one hand, this fits the nature of the investor and company reports, providing first-hand information about the targets' business and performance or detailed (semi-)professional analyses. On the other hand, the low relevance of the news represents the propensity to summarize information and repurpose content (\textit{news recycling}). Stocktwits are multifaceted, they can provide short analyses, first-hand/breaking news information, or promote the opinion of investors who can have an impact with their decisions. 
\par
Regarding the sentiment distribution, we can see that company reports are mainly positive. We hypothesize this is related to the marketing strategy employed by companies to ensure positive reporting, even when the results are poor. Comparing this with table \ref{table:totalobs}, we can observe that VLKAY (N=41), F (N=32), and DDAIF (N=16) were responsible for the largest amount of the considered company reports. Given the prevailing topic of the \textit{Diesel scandal} concerning Volkswagen, a cut of the profit forecast for Ford, and a change of the chief executive officer (CEO) regarding Daimler, company reports might be seen as an instrument to improve the public perception. 
\par
To a lesser extent, the news are also mainly positive. We can further observe that highly relevant news are either very negative or very positive, hence, we can assume these potentially contain important and novel information concerning a target. In contrast, news showing a low positive sentiment and relevance (the majority) are likely dealing with daily, trivial events in a non-polarising way. The number of published news articles, as shown in table \ref{fig:quad_distribution}, also indicates public attention on Ford and Volkswagen (N=96, N=54). 
\par
Overall, the stocktwits distribution contrasts with the remaining sources, it is marked by low relevance with both positive and negative sentiment or neutral sentiment with varying relevance. Moreover, we can observe a rather symmetric distribution of positive and negative sentiment. Stocktwits' short length limits the users' opportunity to thoroughly explain a situation or carefully express a point of view. Contrary to news and reports, StockTwits represent a cross-sectional source. Stocktwits have a broad variety of users with distinct backgrounds and interests, for example, an investor might post a message similar to what one would find in a report, a non-expert user might post similarly to what one would find in a news article or could just express a random thought. This provides reasoning behind the distribution of the stocktwits annotations and the difficulty in its characterisation.
%

%%%%%%%%%%%%%%%%%%%%%%%%%%%%%%%%%%%%%%%%%%%%%%
%%%%%%%%%%%%%%%%% NEW SECTION %%%%%%%%%%%%%%%%
%%%%%%%%%%%%%%%%%%%%%%%%%%%%%%%%%%%%%%%%%%%%%%
\section{Concluding Remarks}
This paper introduces a novel financial corpus for sentiment analysis - the FinLin corpus. It covers four distinct sources representing three data types (reports, news articles, and microblogs). FinLin includes a combined 3,811 texts from investor reports, company reports, news, and stocktwits gathered from 01/July/2018 to 30/September/2018 and targeting a predefined set of entities from the automobile sector. The respective data was annotated for sentiment and relevance by three domain experts and further consolidated by a fourth expert, achieving an overall inter-annotator agreement of 0.1807 for the relevance and 0.5610 for the sentiment.
\par
FinLin's contemporaneous data covering the same entities provide an important resource for textual sentiment analysis across data sources and types. Further, our corpus also has potential applications in behavioural science as it provides insights on the sentiment's relationship  (\emph{i.e.} how sentiment in one source is affected by or affects sentiment in the remaining). A novelty introduced in FinLin's design was the relevance annotation; this provides additional knowledge on a text's relevance to a company. The relevance can potentially improve sentiment analysis, as suggested by previous related work, and poses itself another research task.
\par
During the development of FinLin we encountered a few challenges. The first difficulty relates to the identification of the \textit{point of view}, as well as the difference between \textit{subjective sentiment} and \textit{factual sentiment}. The second challenge was posed by the difficulty for the relevance annotation. While differentiating between relevant and non-relevant was feasible, the challenge is posed by the correct and subjective interpretation of the magnitude of a text's relevance for a given company. As this depends on the annotators' interpretation as well as their background, relevance annotations can exhibit a high variation even among experts. Lastly, the nature of the data source also posed challenges, especially stocktwits for which the limited length can potentially impede the correct interpretation of the text.
\par
Overall, FinLin aims to complement the current knowledge by providing a novel and publicly available financial sentiment dataset and foster research on the topic of financial sentiment analysis.
%%%%%%%%%%%%%%%%%%%%%%%%%%%%%%%%%%%%%%%%%%%%%%
%%%%%%%%%%%%%%%%% NEW SECTION %%%%%%%%%%%%%%%%
%%%%%%%%%%%%%%%%%%%%%%%%%%%%%%%%%%%%%%%%%%%%%%
\section*{Funding}
This publication has emanated from research conducted with the financial support of Science Foundation Ireland (SFI) under Grant Number SFI/12/RC/2289\_P2, co-funded by the European Regional Development Fund.

\bibliographystyle{plainnat}
\bibliography{References.bib}

\end{document}